\documentclass[10pt,letterpaper,twocolumn]{article}

\usepackage[margin=1.9cm]{geometry}

\usepackage{hyperref}
\usepackage{graphicx}
\usepackage{caption}
\usepackage{subcaption}

\begin{document}

\twocolumn[

\title{CodeReef: an open platform for portable MLOps,\\
reusable automation actions and reproducible benchmarking}
\author{Grigori Fursin, Herv\'{e} Guillou, Nicolas Essayan \\
\\
\href{https://CodeReef.ai}{CodeReef.ai}\\
\\
The \href{https://mlops-systems.github.io/}{Workshop on MLOps Systems} at \href{https://mlsys.org/Conferences/2020}{MLSys'20}\\
\\
A live interactive demo: \href{https://CodeReef.ai/demo}{CodeReef.ai/demo}\\
}

\date{January, 2020}

\vskip 0.2in


\maketitle

\begin{abstract}

We present CodeReef - an open platform to share all the components necessary
to enable cross-platform MLOps (MLSysOps), i.e. automating the deployment 
of ML models across diverse systems in the most efficient way.
We also introduce the CodeReef solution - a way to package and share models 
as non-virtualized, portable, customizable and reproducible archive files.
Such ML packages include JSON meta description of models with all dependencies, 
Python APIs, CLI actions and portable workflows necessary to automatically build, benchmark, test 
and customize models across diverse platforms, AI frameworks, libraries, compilers and datasets.

We demonstrate several CodeReef solutions to automatically build, run and measure object detection 
based on SSD-Mobilenets, TensorFlow and COCO dataset
from the latest MLPerf inference benchmark across a wide range of platforms 
from Raspberry Pi, Android phones and IoT devices to data centers.
Our long-term goal is to help researchers share their new techniques 
as production-ready packages along with research papers 
to participate in collaborative and reproducible benchmarking,
compare the different ML/software/hardware stacks and select the most efficient ones on a Pareto frontier
using online CodeReef dashboards.

\end{abstract}

\vspace{0.3cm}


\vspace{0.3cm}

]

\section{Introduction}

\begin{figure*}[ht]
  \centering
  \includegraphics[width=1.0\textwidth]{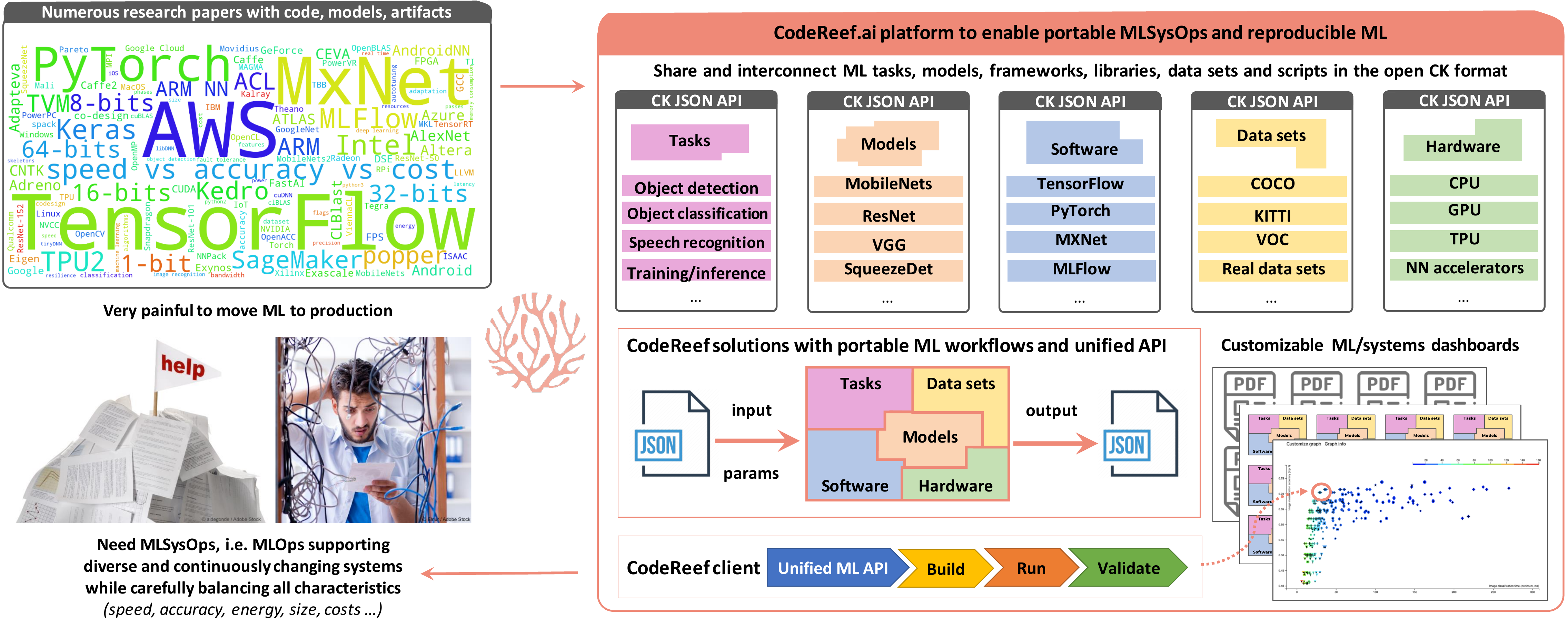}
  \caption{CodeReef.ai: an open platform to keep track of AI/ML/systems research with portable workflows and reproducible crowd-benchmarking}
  \label{fig:codereef-concept}
\end{figure*}

While helping different organizations to deploy machine learning models in production during past 10 years we have noticed that finding the relevant code and training models is only a tip of the MLOps iceberg.
The major challenge afterwards is to figure out how to integrate such models with complex production systems and run them in the most reliable and efficient way across rapidly evolving software, heterogeneous hardware and legacy platforms with continuously changing interfaces and data formats while balancing multiple characteristics including speed, latency, accuracy, memory size, power consumption, reliability and costs.

Different tools were introduced in the past few years to cope with this very tedious, ad-hoc and error-prone process:

\begin{itemize} 

\item
 ML workflow frameworks such as MLFlow~\cite{Zaharia2018AcceleratingTM}, 
 Kedro~\cite{kedro} and Amazon SageMaker~\cite{sagemaker} help to abstract and automate 
 high-level ML operations. However unless used inside AWS or DataBricks cloud they still have a limited support for the complex
 system integration and optimization particularly when targeting embedded devices and IoT - the last mile of MLOps.
 
\item
 ML benchmarking initiatives such as MLPerf~\cite{reddi2019mlperf}, 
 MLModelScope~\cite{li2019acrossstack} and Deep500~\cite{deep500}
 attempt to standardize benchmarking and co-design of models and systems.
 However integration with complex systems and adaptation to continuously 
 changing user environments and data is out of their scope.
 
\item
 Package managers such as Spack~\cite{spack} and EasyBuild~\cite{easybuild}
 are very useful to rebuild the whole environment with fixed software versions.
 However adaptation to existing environment, native cross-compilation 
 and support for non-software packages (models, data sets) is still in progress.

 \item
 Docker, Kubernetes and other container-based technology is very useful to prepare and share stable software releases. However, it hides all the software chaos rather than solving it, has some performance overheads and requires enormous amount of space, have a very poor support for embedded devices and do not help to integrate models with native environment and user data.

 \item
 Collective Knowledge (CK) was introduced as a portable and modular workflow framework to address above issues and bridge the gap between high-level ML operations and systems~\cite{ck-date16,ck2}.
 While it helps companies to automate ML benchmarking and move ML models to production~\cite{GM} we also noticed two major limitations during its practical use: 
  
\begin{itemize} 

\item The distributed nature of the CK technology, the lack of a centralized place to keep all CK components and the lack of convenient GUIs makes it very challenging to keep track of all contributions from the community, add new components, assemble workflows, automatically test them across diverse platforms, and connect them with legacy systems.

\item The concept of backward compatibility of CK APIs and the lack of versioning similar to Java made it very challenging to keep stable and bug-free workflows in real life - a bug in a CK component from one GitHub project can easily break dependent ML workflows in another GitHub project.

\end{itemize} 

\end{itemize}

All these problems motivated us to develop CodeReef (Figure~\ref{fig:codereef-concept}) - an open web-based platform to aggregate, version and test all CK components and portable CK workflows necessary to enable portable MLOps with the automated deployment of ML models in production across diverse systems from IoT to data centers in the most efficient way (MLSysOps).
Importantly, the CodeReef concept is to be non-intrusive and complement, abstract and interconnect all existing tools including MLFlow, SageMaker, Kedro, Spack, EasyBuild, MLPerf, Docker and Kubernetes while making them more adaptive and system aware rather than replacing or rewriting them.
We work with the community to keep track of the state-of-the-art AI, ML and systems research using automated and portable workflows with a common API (compile, run, benchmark, validate) to reproduce and compare results (execution time, accuracy, latency, energy usage, memory consumption and so on) during Artifact Evaluation at systems and ML conferences~\cite{AE}.

\section{CodeReef platform overview}

The CodeReef platform is inspired by GitHub and PyPI: we see it as a collaborative platform to share reusable ML/systems components, portable ML workflows, reproducible ML solutions and associated ML results.
We use open-source technology with open standards instead of proprietary technology to let users exchange all the basic blocks (components) needed to enable portable MLOps without locking them in, assemble portable ML workflows and benchmark their performance using customizable dashboards.

We decided to use the open Collective Knowledge format~\cite{ck2} to share the components and workflows because it is a proven open-source technology used in many academic and industrial projects during the past 4 years~\cite{ck-projects}.
For example, the authors of 18 research papers from different systems and ML conferences used CK to share their research artifacts and workflows~\cite{ck-ae}.

CK helps to convert ad-hoc research projects into a file-based database of reusable components~\cite{ck-modules} (code, data, models, pre-/post-processing scripts, experimental results, R\&D automation actions~\cite{ck-actions}, best research practices to reproduce results, and live papers) with unified Python APIs, CLI-based actions, JSON meta information and JSON input/output.
CK also features plugins to automatically detect different software, models and datasets on a user machine or install/cross-compile the missing ones while supporting different operating systems (Linux, Windows, MacOS, Android) and hardware (Nvidia, Arm, Intel, AMD ...).

Such approach allows researchers to create and share flexible APIs with JSON input/output for different AI/ML frameworks, libraries, compilers, models and datasets, connect them together into unified workflows instead of hardwired scripts, and make them portable~\cite{ck-portable-workflows} using the automatic software detection plugins~\cite{ck-soft-plugins} and meta-packages~\cite{ck-meta-packages}.

We implemented CodeReef platform with an open-source CodeReef client~\cite{codereef-client} to solve two major drawbacks of CK: it is now possible to publish CK components with different versions on a platform, and to download stable versions with all dependencies for a given workflow.
We also provided an API in the CodeReef client to init, run and validate the so-called CodeReef solutions with ML models based on JSON manifest describing all CK dependencies and installation/compilation recipes for different target platforms.

We believe that combining CodeReef and CK can help users to implement and share portable ML workflows assembled from stable and versioned CK components, keep track of all the information flow within such workflows, expose configuration and optimization parameters from different tools and models and change them via JSON input files, combine shared and user code and data, monitor system behavior, retarget ML models with all the necessary software to different platforms from IoT to cloud, use them inside containers, integrate them with legacy systems and reproduce all exposed characteristics - all the pillars of practical MLOps.

\section{CodeReef demo: automating, sharing and reproducing MLPerf inference benchmarks}

A prototype of our platform is now available at \href{https://CodeReef.ai/portal}{CodeReef.ai/portal} and we started working with the community and different organizations to share ML models from research papers and standard benchmarks as portable CodeReef solutions, collaboratively benchmark them across diverse platforms and reproduce performance numbers~\cite{codereef-solutions}.

For example, as a part of the MLPerf benchmarking consortium~\cite{MLP}, we want to automate the manual and tedious process of submitting MLPerf inference benchmark results and share MLPerf models as portable CodeReef solutions. 
Since dividiti, one of the MLPerf submitters, already used open-source CK workflows to submit MLPerf inference results~\cite{mlperf-inference-results}, we demonstrate how to convert them into stable CodeReef solutions with portable ML models, share them with the community, connect them with the CodeReef dashboard and crowdsource benchmarking across diverse platforms provided by volunteers similar to SETI@home at~\href{https://codereef.ai/demo}{CodeReef.ai/demo}.

This live and interactive demo shows how to use the CodeReef solution with a unified API to automatically build, run and validate object detection based on SSD-Mobilenets, TensorFlow and COCO dataset on Raspberry Pi, Android phones, laptops, desktops and data centers.
When preparing this solution we had to manually create a JSON file (we plan to provide a GUI to simplify this process) describing all the dependencies on CK components and APIs to automate the following tasks:
\begin{itemize} 
 \item prepare a Python virtual environment (can be skipped for the native installation),
 \item download and install the Coco dataset (50 or 5000 images),
 \item detect C++ compilers or Python interpreters needed for object detection,
 \item install Tensorflow framework with a specified version for a given target machine,
 \item download and install the SSD-MobileNet model compatible with selected Tensorflow,
 \item manage installation of all other dependencies and libraries,
 \item compile object detection for a given machine and prepare pre/post-processing scripts.
\end{itemize}

We published this solution at the CodeReef platform using the open-source CodeReef client~\cite{codereef-client} to let the users download, initialize, run it and participate in crowd-benchmarking using their machines as follows:
\begin{enumerate}
    \item install CodeReef client from PyPi using:
    
    \textit{pip install codereef}
    
    \item download and install the solution on a given machine (example for Linux): 
    
    \textit{cr init demo-obj-detection-coco-tf-cpu-benchmark-linux-portable-workflows}
    
    \item run the solution on a given machine:
    
    \textit{cr benchmark demo-obj-detection-coco-tf-cpu-benchmark-linux-portable-workflows}
\end{enumerate}

The users can then see all their measurements (speed, latency, accuracy) and compare them against the official MLPerf results using the CodeReef dashboard for this solution at~\href{https://codereef.ai/portal/c/cr-result/sota-mlperf-object-detection-v0.5-crowd-benchmarking}{codereef.ai/portal/c/cr-result/sota-mlperf-object-detection-v0.5-crowd-benchmarking}.

After validating this solution on a given platform, the users can retarget it to other devices and operating systems such as MacOS, Windows with Docker, Android phones, servers with CUDA-enabled GPUs and so on.

We also demonstrate how the unified API of CodeReef solutions can help to integrate ML models with legacy systems and applications by integrating the above workflow with the live object detection from the web cam available at
\href{https://codereef.ai/portal/c/cr-solution/demo-obj-detection-coco-tf-cpu-webcam-linux-azure}{CodeReef.ai/portal/c/cr-solution/demo-obj-detection-coco-tf-cpu-webcam-linux-azure}.

\section{Conclusions}

We presented CodeReef - an open platform to share ML models and related artifacts from research papers and benchmarks as non-virtualized, portable, customizable and reusable open-source packages while automating their benchmarking and deployment in production across diverse platforms and datasets. 
Since we are only at the beginning of this long-term community project, we collaborate with the community, companies, ACM, MLPerf and several systems and machine learning conferences to improve the CodeReef platform and support their reproducibility initiatives, automate ML benchmarks and enable portable MLOps based on real-world use cases.

\bibliographystyle{plain}
\bibliography{codereef}

\end{document}